\documentclass[a4paper,12pt]{article}

\usepackage[utf8]{inputenc}
\usepackage[T1]{fontenc}
\usepackage{fouriernc}
\usepackage[english]{babel}
\usepackage{algorithm,algorithmic,array,setspace}
\usepackage{amsthm,amssymb,amsfonts,amsmath,color,mathrsfs,graphicx,float,url,breakurl,bbm,caption,subfig,placeins,xspace,eucal}
\usepackage[authoryear,round]{natbib}

\usepackage[pagebackref = true]{hyperref}
\hypersetup{
  colorlinks = true,
  urlcolor = blue, 
  linkcolor = blue,
  citecolor = blue,
  pdftitle = {An Oracle Inequality for Quasi-Bayesian Non-Negative Matrix Factorization - \today},
  pdfauthor = {Pierre Alquier and Benjamin Guedj},
  pdfsubject = {}
}

\def\E{\mathbb{E}}

\newtheorem{thm}{Theorem}
\newtheorem{dfn}{Definition}
\newtheorem{lemma}{Lemma}[section]

\newtheorem{coro}{Corollary}
\newtheorem{rmk}{Remark}

\newtheoremstyle{exampstyle}
  {\topsep 3pt} 
  {\topsep 3pt} 
  {\itshape} 
  {} 
  {\bfseries} 
  {.} 
  {.5em} 
  {\thmname{#1}\thmnumber{#2}} 
\theoremstyle{exampstyle}
\newtheorem{cond}{C}

\def\Cnospace~{C{}}

\addto\extrasenglish{%

}

\allowdisplaybreaks
\graphicspath{{fig/}}

\title{An Oracle Inequality for Quasi-Bayesian Non-Negative Matrix Factorization}
\author{Pierre Alquier\footnote{CREST, ENSAE, Université Paris Saclay, \href{mailto:pierre.alquier@ensae.fr}{pierre.alquier@ensae.fr}. This author
gratefully acknowledges financial support from the research programme {\it New Challenges for New Data} from LCL and GENES, hosted by the {\it Fondation du Risque}, from Labex ECODEC (ANR - 11-LABEX-0047) and from Labex CEMPI  (ANR-11-LABX-0007-01).}\, \& Benjamin Guedj\footnote{Modal project-team, Inria Lille - Nord Europe research center, \href{mailto:benjamin.guedj@inria.fr}{benjamin.guedj@inria.fr}.}}
\date{\today}

\begin{document}
\maketitle

\begin{abstract}
\noindent {\bf This is the corrected version of a paper that was published as}

\noindent {\it P. Alquier, B. Guedj, An Oracle Inequality for Quasi-Bayesian Non-negative Matrix Factorization, Mathematical Methods of Statistics, 2017, vol. 26, no. 1, pp. 55-67.}

\noindent {\bf Since then, a mistake was found in the proofs. We fixed the mistake at the price of a slightly different logarithmic term in the bound.~\footnote{We thank Arnak Dalalyan (ENSAE) who found the mistake.}}

\noindent The aim of this paper is to provide some theoretical understanding of
quasi-Bayesian aggregation methods non-negative matrix factorization. We derive
an oracle inequality for an aggregated estimator.
This result holds for a very general class of prior distributions and
shows how the prior affects the rate of convergence.
\end{abstract}


\section{Introduction}

Non-negative matrix factorization (NMF) is a set of algorithms in
high-dimensional
data analysis which aims at factorizing a large matrix $M$ with non-negative entries. If $M$ is an $m_1 \times m_2$ matrix, NMF consists in decomposing it as a product of two matrices of smaller dimensions: $M\simeq UV^T $ where $U$
is $m_1 \times K$, $V$ is $m_2 \times K$, $K\ll m_1 \wedge m_2$ and both
$U$ and $V$ have non-negative entries. Interpreting the columns  $M_{\cdot,j}$
of $M$ as (non-negative) signals, NMF amounts to decompose (exactly, or approximately) each signal as a combination
of the ``elementary'' signals $U_{\cdot,1},\dots,U_{\cdot,K}$:
\begin{equation}
\label{eq-dictionary}
M_{\cdot,j} \simeq \sum_{\ell =1}^K V_{j,\ell} U_{\cdot,\ell}  .
\end{equation}

Since the seminal paper from \cite{LS1999}, NMF was successfully applied to
various fields such as image processing and face
classification \citep{guillamet2002classifying},
separation of sources in audio and video processing \citep{ozerov2010multichannel},
collaborative filtering and recommender systems on the Web \citep{koren2009matrix},
document clustering \citep{xu2003document,shahnaz2006document}, medical image processing \citep{AGT2014} or
topics extraction in texts \citep{paisley2015bayesian}.
In all these applications, it has been pointed out that NMF provides a decomposition which is usually interpretable. \cite{donoho2003does} have given a theoretical foundation to this interpretatibility by exhibiting conditions under which the decomposition $M\simeq UV^T$ is unique. However, let us stress that even when
this is not the case, the results provided by NMF are still sensibly interpreted by practitioners.
\medskip

Since a prior knowledge on the shape and/or magnitude of the signal is available in many settings, Bayesian tools have extensively been used for (general) matrix factorization \citep{CV2004,lim2007variational,salakhutdinov2008bayesian,
lawrence2009non,zhou2010nonparametric} and have been adapted for the Bayesian NMF problem \citep[][among others]{moussaoui2006separation,cemgil2009bayesian,fevotte2009nonnegative,schmidt2009bayesian,tan2009automatic,zhong2009reversible}.
\medskip

The aim of this paper is to provide some theoretical analysis
on the performance of an aggregation method for NMF inspired by the aforementioned Bayesian works. We propose a quasi-Bayesian estimator for NMF. By quasi-Bayesian, we mean that the construction of the estimator relies on a prior distribution $\pi$, however, it does not rely on any parametric assumptions - that is, the likelihood used to build the estimator does not have to be well-specified (it is usually referred to as a quasi-likelihood).
The use of quasi-likelihoods in Bayesian
estimation is advocated by \cite{bissiri2013general} using decision-theoretic arguments. This methodology is also popular in machine
learning, and various authors developed a theoretical framework to analyze it~\citep[][this is known as the PAC-Bayesian theory]{STW,McA,catoni2003pac,catoni2004statistical,MR2483528}. It is also related to recent works on exponentially weigthed aggregation in statistics~\cite{dalalyan2008aggregation,golubev2014concentration}.
Using these theoretical tools, we derive an oracle inequality for our quasi-Bayesian estimator.
The message of this theoretical bound is that our procedure is able to adapt to the unknown rank of $M$ under very general assumptions for the noise.
\medskip

The paper is organized as follows. Notation for the NMF framework and
the definition of our quasi-Bayesian estimator are given in \autoref{section_notation}.
The oracle inequality, which is our main contribution, is given in \autoref{section_theorem} and its
proof is postponed to \autoref{sectionproofs}. The computation of our estimator being completely similar to the computation of a (proper) Bayesian estimator, we end the paper with a short discussion and references to state-of-the-art computational methods for Bayesian NMF in \autoref{section_gibbs}.

\section{Notation}
\label{section_notation}

For any $p\times q$ matrix $A$  we denote by $A_{i,j}$ its
$(i,j)$-th entry, $A_{i,\cdot}$ its $i$-th row and $A_{\cdot,j}$
its $j$-th column. For any $p\times q$ matrix $B$ we define
$$ \left<A,B\right>_F = {\rm Tr}(AB^\top) = \sum_{i=1}^p
\sum_{j=1}^q A_{i,j} B_{i,j} .$$
We define the Frobenius norm $ \|A\|_F$ of $A$ by
$ \|A\|_F^2 = \left<A,A\right>_F$. Let $A_{-i,\cdot}$ denote the
matrix $A$ where the $i$-th column is removed. In the same way, for
a vector $v\in\mathbb{R}^p$, $v_{-i}\in\mathbb{R}^{p-1}$ is the vector
$v$ with its $i$-th coordinate removed. Finally, let $ {\rm Diag}(v)$ denote
the $p\times p$ diagonal matrix given by $ [{\rm Diag}(v)]_{i,i} = v_i$.

\subsection{Model}

The object of interest is an $m_1 \times m_2$ target matrix $M$
possibly polluted with some noise $\mathcal{E}$. So we actually
observe
\begin{equation}
\label{modele}
Y = M + \mathcal{E},
\end{equation}
and we assume that $\mathcal{E}$ is random with $\E(\mathcal{E})
=0$. The objective is to approximate $M$ by a
matrix $UV^T$ where $U$
is $m_1 \times K$, $V$ is $m_2 \times K$ for some $K\ll m_1 \wedge m_2$, and where $U$, $V$ and $M$ all
have non-negative entries.
Note that, under~\eqref{modele}, depending on the distribution of $\mathcal{E}$,
$Y$ might have some negative entries (the non-negativity assumption is on $M$ rather than on $Y$).
Our theoretical analysis only requires the following assumption on $\mathcal{E}$.
\begin{cond}
 \label{condition_noise}
 The entries $\mathcal{E}_{i,j}$ of $\mathcal{E}$ are i.i.d. with
 $\mathbb{E}(\varepsilon_{i,j})=0$.
With the notation $m(x)=\mathbb{E}[\varepsilon_{i,j}
  \mathbf{1}_{(\varepsilon_{i,j}\leq x)}]$ and $ F(x) =
  \mathbb{P}(\varepsilon_{i,j}\leq x)$, assume that
  there exists a non-negative and bounded function $g$ with $\|g\|_{\infty} \leq 1$
 and
 \begin{equation}
 \label{eq_c_n}
 \int_{u}^{v} m(x) {\rm d}x = \int_{u}^{v} g(x) {\rm d}F(x) .
 \end{equation}
\end{cond}

First, note that if \eqref{eq_c_n} is satisfied for a function $g$ with
$\|g\|_{\infty} = \sigma^2 >1$, we can replace \eqref{modele} by the normalized model $Y/\sigma = M/\sigma + \varepsilon/\sigma$ for which \autoref{condition_noise} is satisfied.
The introduction of this rather involved condition is due to the technical
analysis of our estimator which is based on \autoref{thm_dt} in \autoref{sectionproofs}. \autoref{thm_dt} has first been proved by \cite{DT2007} using Stein's formula with a Gaussian noise. However, \cite{dalalyan2008aggregation} have shown that \autoref{condition_noise} is actually sufficient to prove \autoref{thm_dt}. For the sake of understanding, note that \autoref{eq_c_n} is fulfilled when the noise is Gaussian ($\varepsilon_{i,j}\sim\mathcal{N}(0,\sigma^2)$ with $\|g\|_{\infty}=\sigma^2$) or uniform ($\varepsilon_{i,j}\sim\mathcal{U}[-b,b]$ with $\|g\|_{\infty}=b^2/2$).

\subsection{Prior}

We are going to define a prior $\pi(U,V)$, where $U$ is $m_1\times K$ and $V$ is $m_2 \times K$, for a fixed $K$. Regarding the choice of $K$, we prove in \autoref{section_theorem} that our quasi-Bayesian estimator is adaptive, in the sense that if $K$ is chosen much larger than the actual rank of $M$, the prior will put very little mass on many columns of $U$ and $V$, automatically shrinking them to $0$. This seems to advocate for setting a large $K$ prior to the analysis, say $K=m_1 \wedge m_2$. However, keep in mind that the algorithms discussed below have a computational cost growing with $K$. Anyhow, the following theoretical analysis only requires $2 \leq K \leq m_1 \wedge m_2$.\medskip

With respect to
the Lebesgue measure
on $\mathbb{R}_+$, let us fix a density $f$ such that $$ S_f := 1 \vee \int_0^{\infty} x^2 f(x) {\rm d}x < + \infty.  $$ For any $a,x>0$, let
$$
g_{a}(x) := \frac{1}{a}f\left(\frac{x}{a}\right).
$$
We define the prior on $U$ and $V$ by
$$ U_{i,\ell}, V_{i,\ell} \text{ indep. }
   \sim g_{\gamma_{\ell}}(\cdot) $$
where
$$ \gamma_\ell \text{ indep. } \sim h(\cdot) $$
and $h$ is a density on $\mathbb{R}_+$.
With the notation $\gamma=(\gamma_1,\dots,\gamma_K)$, define $\pi$ by
\begin{equation}
\label{def_prior}
\pi(U,V,\gamma)
      = \prod_{\ell=1}^K \left(\prod_{i=1}^{m_1} 
      g_{\gamma_{\ell}}(U_{i,\ell})\right)
      \left(\prod_{j=1}^{m_2}
      g_{\gamma_{\ell}}(V_{j,\ell})\right)
      h(\gamma_\ell)
\end{equation}
and
$$ \pi(U,V) = \int_{\mathbb{R}_+^K}  \pi(U,V,\gamma) {\rm d}\gamma .$$
The idea behind this prior is that under $h$, many $\gamma_{\ell}$
should be small and lead to non-significant columns $U_{\cdot,\ell}$
and $V_{\cdot,\ell}$. In order to do so, we must assume that a non-negligible
proportion of the mass of $h$ is located around $0$. On the other hand, a non-neglibible probability must be assigned to significant values. This is the meaning of the following assumption.

\begin{cond}
 \label{condition_zero_h}
 There exist constants $0<\alpha<1$, $\beta \geq 0$ and $\delta>0$ such that
 for any $0<\varepsilon\leq \frac{1}{2\sqrt{2} S_f} $,
 $$ \int_{0}^{\varepsilon}h(x) {\rm d}x \geq \alpha
     \varepsilon^\beta \text{ and }
     \int_{1}^{2}h(x) {\rm d}x \geq \delta. $$
\end{cond}

Finally, the following assumption on $f$ is required to prove
our main result.

\begin{cond}
 \label{condition_lowerb_f}
 There exist a non-increasing density $\widetilde{f}$ w.r.t. the Lebesgue measure
 on $\mathbb{R}_+$ and a constant $\mathcal{C}_f>0$  such that
 for any $x>0$,
 $$ f(x) \geq \mathcal{C}_f \widetilde{f}(x). $$
\end{cond}

As shown in \autoref{main_thm}, the heavier the tails of $\widetilde{f}(x)$, the better 
the performance of Bayesian NMF.
\medskip

Note that the general form of~\eqref{def_prior} encompasses as
special cases almost all the priors used  in the papers mentioned in the introduction.
We end this subsection with classical examples of functions $f$ and $h$.
Regarding $f$:
\begin{enumerate}
 \item Exponential prior $f(x) = \exp(-x)$ with $\widetilde{f}=f$,
$\mathcal{C}_f=1$ and $S_f=2$. This is the choice made by~\cite{schmidt2009bayesian}.
 A generalization of the exponential prior is the gamma prior used in~\cite{cemgil2009bayesian}.
 \item Truncated Gaussian prior $f(x) \propto \exp(2ax-x^2) $ with $a\in\mathbb{R}$.
 \item Heavy-tailed prior $f(x) \propto \frac{1}{(1+x)^{\zeta}}$ with $\zeta>1$. This choice is inspired by \cite{dalalyan2008aggregation} and leads to better theoretical properties.
\end{enumerate}

Regarding $h$:
\begin{enumerate}
\item The uniform distribution on $[0,2]$ obviously satisfies \autoref{condition_zero_h}
with $\alpha=1/2$, $\beta = 1$ and $\delta=1/2$.
\item The inverse gamma prior $h(x)=\frac{b^a}{\Gamma(a)}
\frac{1}{x^{a+1}} \exp\left(-\frac{b}{x}\right) $
is classical in the literature for computational reasons \citep[see for example][]{salakhutdinov2008bayesian,alquier2013bayesian}, but note that it does not satisfy  \autoref{condition_zero_h}.
\item \cite{alquier2014bayesian} discuss the $\Gamma(a,b)$ choice for $a,b>0$:
both gamma and inverse gamma lead to explicit conditional posteriors for $\gamma$ (under a restriction on $a$ in the second case), but the gamma distribution led to better numerical performances. When $h$ is the density of the $\Gamma(a,b)$, \autoref{condition_zero_h} is satisfied with $\beta = a$ and $\alpha= b^{a} \exp[-b/(2\sqrt{2} S_f)] / \Gamma(a+1) $ and $\delta = \int_{1}^{2} b^a x^{a-1} \exp(-bx) {\rm d}x / \Gamma(a) $.
\end{enumerate}

\subsection{Quasi-posterior and estimator}

We define the quasi-likelihood as
$$ \widehat{L}(U,V) = \exp\left[-\lambda \|Y-UV^\top\|_F^2 \right] $$
for some fixed parameter $\lambda>0$.
Note that under the assumption that $\varepsilon_{i,j} \sim \mathcal{N}(0,
1/(2\lambda))$, this would be the actual likelihood
up to a multiplicative constant. As already pointed out, the use
of quasi-likelihoods to define quasi-posteriors is becoming rather popular
in Bayesian statistics and machine
learning literatures. Here, the Frobenius norm is to be seen as a fitting criterion rather than
as a ground truth. Note that other criterion were used in the literature: the Poisson likelihood~\citep{LS1999},
or the Itakura-Saito divergence~\citep{fevotte2009nonnegative}.
\medskip

\begin{dfn}
We define the quasi-posterior as
\begin{align*}
\widehat{\rho}_{\lambda}(U,V,\gamma)
&= \frac{1}{Z} \widehat{L}(U,V) 
 \pi(U,V,\gamma) \\
 &= \frac{1}{Z} \exp\left[-\lambda \|Y-UV^\top\|_F^2 \right]
 \pi(U,V,\gamma),
\end{align*}
where
$$
Z := \int \exp\left[-\lambda \|Y-UV^\top\|_F^2 \right]
 \pi(U,V,\gamma)  {\rm d}(U,V,\gamma)
$$
is a normalization constant. The posterior mean will be denoted by
$$
\widehat{M}_{\lambda}
=
 \int UV^T \widehat{\rho}_{\lambda}(U,V,\gamma) {\rm d}(U,V,\gamma).
$$
\end{dfn}

\autoref{section_theorem} is devoted to the study the theoretical properties of $
\widehat{M}_{\lambda}$.
A short discussion on the implementation will be provided in~\autoref{section_gibbs}.

\section{An oracle inequality}
\label{section_theorem}

Most likely, the rank of $M$ is unknown in practice. So, as
recommended above, we usually choose $K$ much larger than the
expected order for the rank, with the hope that many columns of $U$
and $V$ will be shrinked to $0$. The following set of matrices is introduced to formalize this idea.
For any $r\in \{1,\dots,K\}$, let $ \mathcal{M}_r $ be the set of pairs of matrices
$(U^0,V^0)$ with non-negative entries such that
$$ U^0 = \left(
\begin{array}{c c c c c c}
 U^0_{11} & \dots & U^0_{1r} & 0 & \dots & 0 \\
 \vdots & \ddots & \vdots & \vdots & \ddots & \vdots \\
 U^0_{m_11} & \dots & U^0_{m_1r} & 0 & \dots & 0
\end{array}
\right),
V^0 = \left(
\begin{array}{c c c c c c}
 V^0_{11} & \dots & V^0_{1r} & 0 & \dots & 0 \\
 \vdots & \ddots & \vdots & \vdots & \ddots & \vdots \\
 V^0_{m_2 1} & \dots & V^0_{m_2 r} & 0 & \dots & 0
\end{array}
\right).
$$
We also define $\mathcal{M}_r(L)$ as the set of matrices
$(U^0,V^0)\in\mathcal{M}_r$ such that, for any $(i,j,\ell)$,
$U_{i,\ell}^0,V_{j,\ell}^0\leq L$.
\medskip

We are now in a position to state our main theorem, in the form of the following oracle inequality.

\begin{thm}
 \label{main_thm}
Fix $\lambda = 1/4$.
Under assumptions
\autoref{condition_noise}, \autoref{condition_zero_h} and \autoref{condition_lowerb_f},
\begin{multline*}
 \E\left(
 \|\widehat{M}_{\lambda} - M\|_F^2
 \right)
 \leq
 \inf_{1 \leq r \leq K}\
 \inf_{(U^0,V^0)\in\mathcal{M}_r}\
 \Biggl\{
 \|U^0 V^{0\top}-M\|_F^2
 \\
  +  \mathcal{R}(r,m_1,m_2,M,U^0,V^0,\beta,\alpha,\delta,K,S_f,\tilde{f})  \Biggr\}.
\end{multline*}
  where
\begin{multline*}
\mathcal{R}(r,m_1,m_2,M,U^0,V^0,\beta,\alpha,K,S_f,\tilde{f})
\\
=  8(m_1 \vee m_2) r
   \log\left(
  \sqrt{2(m_1 \vee m_2)r}
  \right)
  \\
  +  8(m_1 \vee m_2) r
   \log\left(
   \frac{\left[1+ \|U^0\|_F + \| V^{0} \|_F+ 2 \|  U^0  V^{0\top} - M \|_F\right]^2}{ \mathcal{C}_f }
  \right)
 \\
 + 4\sum_{\substack{1 \leq i \leq m_1\\1\leq \ell \leq r}}
          \log\left(\frac{1}{\widetilde{f}(U_{i\ell}^0 +1)}
          \right)
 + 4\sum_{\substack{1 \leq j \leq m_2\\1\leq \ell \leq r}}
          \log\left(\frac{1}{\widetilde{f}(V_{j\ell}^0 +1)}
          \right)
 \\
 + 4\beta K    \log\left(
   \left[1+ \|U^0\|_F + \| V^{0} \|_F+ 2 \|  U^0  V^{0\top} - M \|_F\right]^2
  \right)
  \\
 + 4   \beta K \log\left(2 S_f \sqrt{2  K (m_1 \vee m_2)}\right)
   \\
    + r\left[4\log\left(\frac{1}{\delta}\right)\right] + 4 K \log\left(\frac{1}{\alpha}\right) + 4 \log(4) +1.
\end{multline*}
\end{thm}

We remind the reader that the proof is given in \autoref{sectionproofs}.
The main message of the theorem is that $\widehat{M}_{\lambda}$ is as close to $M$ as would
be an estimator designed with the actual knowledge of its rank (\emph{i.e.}, $\widehat{M}_\lambda$ is adaptive to $r$), up to remainder
terms. These terms might be difficult to read.
In order to explicit the rate of convergence, we now provide a weaker version, where
we assume that $M=U^0V^{O\top}$ for some $(U^0,V^0)\in\mathcal{M}_r(L)\mathcal{M}_r(L)$; note that the estimator $\widehat{M}_{\lambda}$ still doesn't depend on $L$ nor on $r$.

\begin{coro}
\label{main_thm_cor}
Fix $\lambda = 1/4$.
Under assumptions
\autoref{condition_noise}, \autoref{condition_zero_h} and \autoref{condition_lowerb_f}, and when $M=U^0V^{O\top}$ for some $(U^0,V^0)\in\mathcal{M}_r(L)\mathcal{M}_r(L)$,
\begin{multline*}
 \E\left(
 \|\widehat{M}_{\lambda} - M\|_F^2
 \right)
 \leq
  8(m_1 \vee m_2) r
   \log\left(
   \frac{2(m_1 \vee m_2)r(1+ 2L\sqrt{r(m_1 \vee m_2)} )^2}{\mathcal{C}_f \tilde{f}(L+1)}
  \right)
  \\
 + 4   \beta K \log\left(2 S_f \sqrt{2  K (m_1 \vee m_2)} (1+ 2L\sqrt{r(m_1 \vee m_2)} )^2\right)
   \\
    + r\left[4\log\left(\frac{1}{\delta}\right)\right] + 4 K \log\left(\frac{1}{\alpha}\right) + 4 \log(4) +1.
\end{multline*}
\end{coro}
First, note that  when $L^2=\mathcal{O}(1)$, the magnitude
of the error bound is
$$ (m_1 \vee m_2) r \log(m_1 m_2),$$
which is roughly the variance
multiplied by the number of parameters to be estimated in any
$(U^0,V^0)\in\mathcal{M}_r(L)$.
Alternatively, when $M=U^0 V^{0\top}$ only for $(U^0,V^0)\in\mathcal{M}_r(L)$ for a huge $L$,
the $\log$ term in
$$
8(m_1 \vee m_2) r
   \log\left( \frac{(L+1)^2 m_1 m_2 }{\tilde{f}(L+1)}
  \right)
$$
becomes significant.
Indeed, in the case of the truncated Gaussian prior $f(x) \propto \exp(2ax-x^2) $,
the previous quantity is in
$$8  (m_1 \vee m_2) r L^2 \log(L m_1 m_2 )$$
which is terrible for large $L$.
On the contrary, with the heavy-tailed
prior $f(x)\propto (1+x)^{-\zeta}$ \citep[as in][]{dalalyan2008aggregation},
the leading term is
$$ 8  (m_1 \vee m_2) r (\zeta+2)\log(L m_1 m_2) $$
which is way more satisfactory. Still, this prior has not received much attention from practitioners.

\begin{rmk}
 When \eqref{eq_c_n} in \autoref{condition_noise} is satisfied with $\|g\|_{\infty}=\sigma^2 >1$ we already remarked that it is necessary to use the normalized model $Y/\sigma=M/\sigma+\mathcal{E}/\sigma$ in order to apply \autoref{main_thm}. Going back to the original model, we get that, for $\lambda = 1/(4\sigma^2)$,
 \begin{multline*}
 \E\left(
 \|\widehat{M}_{\lambda} - M\|_F^2
 \right)
 \leq
 \inf_{1 \leq r \leq K}\
 \inf_{(U^0,V^0)\in\mathcal{M}_r}\
 \Biggl\{
 \|U^0 V^{0\top}-M\|_F^2
 \\
  + \sigma^2 \mathcal{R}(r,m_1,m_2,M,U^0,V^0,\beta,\alpha,\delta,K,S_f,\tilde{f})  \Biggr\}.
\end{multline*}
\end{rmk}

\section{Algorithms for Bayesian NMF}
\label{section_gibbs}

As the quasi-Bayesian estimator takes the form of a Bayesian estimator in a special model, we can obviously use tools from computational Bayesian statistics to compute it. The method of choice for computing Bayesian estimators for complex models
is Monte-Carlo Markov Chain (MCMC). In the case of Bayesian matrix factorization,
the Gibbs sampler was considered in the literature: see for example~\cite{salakhutdinov2008bayesian},
\cite{alquier2014bayesian} for the general case and~\cite{moussaoui2006separation}, \cite{schmidt2009bayesian}
and \cite{zhong2009reversible} for NMF.
The Gibbs sampler \citep[described in its general form in][for example]{Bishop2006},
is given by \autoref{Gibbs}.

\begin{algorithm}[h]
\caption{Gibbs sampler.}\label{Gibbs}
\begin{description}
\item[Input] $Y$, $\lambda$.
\item[Initialization] $U^{(0)}$, $V^{(0)}$, $\gamma^{(0)}$.
\item[For] $k=1,\dots,N$:
\begin{description}
\item[For] $i=1,\dots,m_1$: draw
      $ U_{i,\cdot}^{(k)} \sim \widehat{\rho}_\lambda (U_{i,\cdot}
          |V^{(k-1)},\gamma^{(k-1)},Y) $.
\item[For] $j=1,\dots,m_2$: draw
      $ V_{j,\cdot}^{(k)} \sim \widehat{\rho}_\lambda (V_{j,\cdot}
          |U^{(k)},\gamma^{(k-1)},Y) $.
\item[For] $\ell=1,\dots,K$: draw
      $ \gamma_{\ell}^{(k)} \sim \widehat{\rho}_\lambda (\gamma_{\ell}|U^{(k)}
          ,V^{(k)},Y) $.
\end{description}
\end{description}
\end{algorithm}

In the aforementioned papers, there are discussions on the choices of $f$ and $h$ that leads to explicit forms for the conditional posteriors of $ U_{i,\cdot}$, $ V_{j,\cdot}$ and $\gamma_{\ell}$, leading to fast algorithms. We refer the reader to these papers for detailed descriptions of the algorithm in this case, and for exhautstive simulations studies.

Optimization methods used for (non-Bayesian) NMF are much faster than the MCMC methods used for Bayesian NMF though: the original multiplicative algorithm \cite{LS1999,lee2001algorithms}, projected gradient descent \citep{lin2007projected,guan2012nenmf}, second order schemes \citep{kim2008fast}, linear progamming~\citep{recht2012factoring}, ADMM~\citep[alternative direction method of multipliers][]{boyd2011distributed,xu2012alternating}, block coordinate descent \cite{xu2013block} among others.

We believe that an efficient implementation of Bayesian and quasi-Bayesian methods will be based on fast optimisation methods, like Variational Bayes (VB) or Expectation-Progapation (EP) methods~\citep{Jordan1999,MacKay2002,Bishop2006}. VB was used for Bayesian matrix factorization~\citep{lim2007variational,alquier2014bayesian}
and more recently in Bayesian NMF~\citep{paisley2015bayesian} with promising results. Still, there is no proof that these algorithms provide valid results. To the best of our knowledge, the first attempt to study the convergence of the VB to the target distribution is studied in~\cite{Alquier2015} for a family of problems, that do not include NMF. We believe that further investigation in this direction is necessary.


\section{Proofs}
\label{sectionproofs}

This section contains the proof to the main theoretical claim of the paper (\autoref{main_thm}).

\subsection{A PAC-Bayesian bound from \cite{dalalyan2008aggregation}}

The analysis of quasi-Bayesian estimators with PAC bounds started with \cite{STW}.
McAllester improved on the initial method and introduced the name ``PAC-Bayesian bounds''
\citep{McA}. Catoni also improved these results to derive
sharp oracle inequalities \citep{catoni2003pac,catoni2004statistical,MR2483528}.
This methods were used in various complex models of statistical
learning~\citep{Guedj2013,alquier2013bayesian,suzuki2014,Mai2015,GR2015,ilaria2015,li2016pac}.
\cite{dalalyan2008aggregation} proved a different
PAC-Bayesian bound based on the idea of unbiased risk
estimation \citep[see][]{leung2006information}.
We first recall its form in the context of matrix factorization.

\begin{thm}
 \label{thm_dt}
 Under \autoref{condition_noise}, as soon as
  $\lambda\leq 1/4$,
 $$
 \E
 \|\widehat{M}_{\lambda} - M\|_F^2
 \leq
 \inf_{\rho}\
 \left\{
 \int \| U V^\top - M \|_F^2 \rho(U,V,\gamma) {\rm d}(U,V,\gamma)
 + \frac{\mathcal{K}(\rho,\pi)}{\lambda}
 \right\},
 $$
 where the infimum is taken over all probability measures $\rho$ absolutely continuous with respect to $\pi$, and $\mathcal{K}(\mu,\nu)$ denotes the Kullback-Leibler divergence between two measures $\mu$ and $\nu$.
\end{thm}

We let the reader check that the proof in \cite{dalalyan2008aggregation},
stated for vectors, is still valid for matrices (also, the result \cite{dalalyan2008aggregation} is actually stated for any $\sigma^2$, we only use the case $\sigma^2=1$).
\medskip

The end of the proof of \autoref{main_thm} is organized as follows. First, we define in \autoref{param} a
parametric family of probability distributions $\rho$:
$$\left\{\rho_{r,U^0,V^0,c} \colon c>0, 1\leq r \leq K, (U^0,V^0)\in\mathcal{M}_r\right\}.$$
We then upper bound the infimum over all $\rho$ by the infimum
over this parametric family. So, we have to calculate, or upper bound
$$
 \int \| U V^\top - M \|_F^2 \rho_{r,U^0,V^0,c}(U,V,\gamma) {\rm d}(U,V,\gamma)
$$
and
$$
\mathcal{K}(\rho_{r,U^0,V^0,c},\pi).
$$
This is done in two lemmas in
\autoref{integral} and \autoref{kullback} respectively.
We finally gather all the pieces together in \autoref{conclusion},
and optimize with respect to $c$.

\subsection{A parametric family of factorizations}
\label{param}

We define, for any $r\in\{1,\dots,K\}$
and any pair of matrices $(U^0,V^0)\in\mathcal{M}_r$,
for any $0<c\leq 1$,
the density
$$ \rho_{r,U^0,V^0,c}(U,V,\gamma)
 = \frac{ \mathbf{1}_{\left\{\|U-U^0\|_F \leq c,
        \|V-V^0\|_F \leq c\right\}}
          \pi(U,V,\gamma)}
        {\pi\left(\left\{\|U-U^0\|_F \leq c,
        \|V-V^0\|_F \leq c \right\}\right) }. $$

\subsection{Upper bound for the integral part}
\label{integral}

\begin{lemma}
We have
 \label{lemma_integral}
\begin{multline*}
\int \| U V^\top - M \|_F^2
\rho_{r,U^0,V^0,c}(U,V,\gamma){\rm d}(U,V,\gamma)
\leq \|  U^0  V^{0\top} - M \|_F^2
+
\\ c \left( 1+ \|U^0\|_F + \| V^{0} \|_F+ 2 \|  U^0  V^{0\top} - M \|_F \right)^2.
\end{multline*}
\end{lemma}

\begin{proof}
We have
\begin{align*}
\int & \| U V^\top - M \|_F^2
\rho_{r,U^0,V^0,c}(U,V,\gamma){\rm d}(U,V,\gamma)
\\
& =
\int  \Biggl( \| U V^\top - U^0  V^{0\top} \|_F^2
 + 2 \left<U V^\top - U^0  V^{0\top}, U^0  V^{0\top} - M\right>_F
 \\
& \quad + \|  U^0  V^{0\top} - M \|_F^2 \Biggr)
\rho_{r,U^0,V^0,c}(U,V,\gamma){\rm d}(U,V,\gamma)
\\
& \leq
\int  \| U V^\top - U^0  V^{0\top} \|_F^2 \rho_{r,U^0,V^0,c}(U,V,\gamma){\rm d}(U,V,\gamma)
 \\
& \quad  + 2 \sqrt{\int \| U V^\top - U^0  V^{0\top} \|_F^2  \rho_{r,U^0,V^0,c}(U,V,\gamma){\rm d}(U,V,\gamma)} \|  U^0  V^{0\top} - M \|_F
 \\
& \quad + \|  U^0  V^{0\top} - M \|_F^2
\end{align*}
Note that $(U,V)$ belonging to the support of $\rho_{r,U^0,V^0,c}$ implies that
\begin{align*}
\|UV^\top -   U^0 V^{0\top} \|_F
& = \| U (V^\top - V^{0\top})
  +  (U-U^0) V^{0\top}\|_F
 \\
 &  \leq
 \| U (V^\top - V^{0\top}) \|_F
  + \|  (U-U^0) V^{0\top} \|_F
  \\
  & \leq
      \|U\|_F \|V - V^{0} \|_F
  + \| U-U^0 \|_F \| V^{0} \|_F 
  \\
  & \leq
        (\|U^0\|_F + c) c
  + c \| V^{0} \|_F 
  \\
  & = c \left( \|U^0\|_F
  + \| V^{0} \|_F +c \right)
\end{align*}
and so
\begin{align*}
\int & \| U V^\top - M \|_F^2
\rho_{r,U^0,V^0,c}(U,V,\gamma){\rm d}(U,V,\gamma)
\\
&  \leq c^2 \left( \|U^0\|_F
  + \| V^{0} \|_F +c \right)^2
 \\
& \quad  + 2 c \left( \|U^0\|_F
  + \| V^{0} \|_F +c \right) \|  U^0  V^{0\top} - M \|_F
 \\
& \quad + \|  U^0  V^{0\top} - M \|_F^2
\\
&  = c \left( \|U^0\|_F
  + \| V^{0} \|_F +c \right) \left[c \left( \|U^0\|_F
  + \| V^{0} \|_F +c \right) + 2 \|  U^0  V^{0\top} - M \|_F \right]
 \\
& \quad + \|  U^0  V^{0\top} - M \|_F^2
\\
&  \leq c \left( \|U^0\|_F
  + \| V^{0} \|_F +1 \right) \left[\left( \|U^0\|_F
  + \| V^{0} \|_F +1\right) + 2 \|  U^0  V^{0\top} - M \|_F \right]
 \\
& \quad + \|  U^0  V^{0\top} - M \|_F^2
\\
&  \leq c  \left[\left( \|U^0\|_F
  + \| V^{0} \|_F +1\right) + 2 \|  U^0  V^{0\top} - M \|_F \right]^2 + \|  U^0  V^{0\top} - M \|_F^2
.
\end{align*}
\end{proof}

\subsection{Upper bound for the Kullback-Leibler divergence}
\label{kullback}

\begin{lemma}
 \label{lemma_kullback}
 Under \autoref{condition_zero_h}
 and \autoref{condition_lowerb_f},
 \begin{multline*}
 \mathcal{K}(\rho_{r,U^0,V^0,c},\pi)
 \leq
 2 (m_1 \vee m_2) r \log\left( \frac{\sqrt{2 (m_1\vee m_2) r}}
                        {c \mathcal{C}_f} \right)
 \\
 + \sum_{\substack{1 \leq i \leq m_1\\1\leq \ell \leq r}}
          \log\left(\frac{1}{ \widetilde{f}(U_{i\ell}^0 +1 )}
          \right)
 + \sum_{\substack{1 \leq j \leq m_2\\1\leq \ell \leq r}}
          \log\left(\frac{1}{ \widetilde{f}(V_{j\ell}^0 +1)}
          \right)
 \\
 + \beta K \log\left(\frac{2 S_f \sqrt{2  K (m_1 \vee m_2)}}
         {c}\right) + K\log\left(\frac{1}{\alpha}\right) + r\log\left(\frac{1}{\delta}\right) + \log(4).
 \end{multline*}
\end{lemma}

\begin{proof}
By definition
\begin{align*}
  \mathcal{K}(\rho_{r,U^0,V^0,c},\pi)
  & = \int \rho_{r,U^0,V^0,c} (U,V,\gamma)
      \log\left(\frac{ \rho_{r,U^0,V^0,c}(U,V,\gamma)}
       {\pi(U,V,\gamma)}\right){\rm d}(U,V,\gamma)
 \\
 & = \log \left(
 \frac{1}{\int
 \mathbf{1}_{\{\|U-U^0\|_F \leq c,
        \|V-V^0\|_F \leq c\}} \pi(U,V,\gamma) {\rm d}(U,V,\gamma)
 }
 \right).
\end{align*}
Then, note that
\begin{align*}
 & \int
 \mathbf{1}_{\{\|U-U^0\|_F \leq c,
        \|V-V^0\|_F \leq c\}} \pi(U,V,\gamma) {\rm d}(U,V,\gamma)
 \\
 & \quad
 = \int \left( \int 
 \mathbf{1}_{\{\|U-U^0\|_F \leq c,
        \|V-V^0\|_F \leq c\}} \pi(U,V|\gamma) {\rm d}(U,V) \right)
         \pi(\gamma) {\rm d}\gamma
  \\
 & \quad
 = \int \underbrace{ \left( \int 
 \mathbf{1}_{\{\|U-U^0\|_F \leq c} \pi(U|\gamma)
          {\rm d}U \right)
          }_{=: I_1(\gamma)}
  \underbrace{  \left( \int 
 \mathbf{1}_{\{\|V-V^0\|_F \leq c} \pi(V|\gamma)
          {\rm d}V \right)
         }_{=: I_2(\gamma)} \pi(\gamma)  {\rm d}\gamma.
\end{align*}
So we have to lower bound $I_1(\gamma)$ and $I_2(\gamma)$. We deal only
with $I_1(\gamma)$, as the method to lower bound $I_2(\gamma)$ is exactly the
same. We define the set $E\subset\mathbb{R}^K$
as
$$ E = \left\{
\gamma \in\mathbb{R}^K: \gamma_1,\dots,\gamma_r \in\left(1,2\right]
\text{ and } \gamma_{r+1},\dots,\gamma_K \in \left(0,
  \frac{c}{2 S_f\sqrt{2K m_1 \vee m_2}}\right]
\right\} .$$
Then
$$
\int I_1(\gamma) I_2(\gamma)
         \pi(\gamma) {\rm d}\gamma
\geq
\int_E I_1(\gamma) I_2(\gamma)
         \pi(\gamma) {\rm d}\gamma
$$
and we focus on a lower-bound for $I_1(\gamma)$
when $\gamma\in E$. 
\begin{align*}
I_1(\gamma)
  & 
  = \pi\left(\sum_{\substack{1 \leq i \leq m_1\\1\leq \ell \leq K}}
        (U_{i,\ell}-U^0_{i,\ell})^2 \leq c^2 \middle\vert\gamma\right)
  \\
  & 
    = \pi\left(\sum_{\substack{1 \leq i \leq m_1\\1\leq \ell \leq r}}
        (U_{i,\ell}-U^0_{i,\ell})^2
        + \sum_{\substack{1 \leq i \leq m_1\\r+1\leq \ell \leq K}}
        U_{i,\ell}^2 \leq c^2 \middle\vert\gamma\right)
   \\
   & 
   \geq    \pi\left(\sum_{\substack{1 \leq i \leq m_1\\r+1\leq \ell \leq K}}
        U_{i,\ell}^2 \leq \frac{c^2}{2} \middle\vert\gamma\right)
        \pi\left(\sum_{\substack{1 \leq i \leq m_1\\1\leq \ell \leq r}}
        (U_{i,\ell}-U^0_{i,\ell})^2 \leq \frac{c^2}{2}\middle\vert\gamma\right)
   \\
     &
   \geq 
   \underbrace{
   \pi\left(\sum_{\substack{1 \leq i \leq m_1\\r+1\leq \ell \leq K}}
      U_{i,\ell}^2 \leq \frac{c^2}{2} \middle\vert\gamma\right) }_{=:I_3(\gamma)}
 \prod_{\substack{1 \leq i \leq m_1\\1\leq \ell \leq r}}
 \pi\left(
        (U_{i,\ell}-U^0_{i,\ell})^2 \leq \frac{c^2}{2m_1 r}
         \middle\vert\gamma\right).
\end{align*}
Now, using Markov's inequality,
\begin{align*}
 1-I_3(\gamma) & =  \pi\left(\sum_{\substack{1 \leq i \leq m_1\\r+1\leq \ell \leq K}}
      U_{i,\ell}^2 \geq \frac{c^2}{2} \middle\vert\gamma\right)
      \\
     & \leq 2\frac{ \E_{\pi} \left(\sum_{\substack{1 \leq i \leq m_1\\r+1\leq \ell \leq K}}
      U_{i,\ell}^2 \middle\vert\gamma\right) } {c^2}
      \\
      & =    2 \frac{ \sum_{\substack{1 \leq i \leq m_1\\r+1\leq \ell \leq K}}
      \gamma_j^2 S_f^2  } {c^2}
      \\
      & \leq \frac{1}{2},
\end{align*}
and as on $E$, for $\ell \geq r+1$, $\gamma_j \leq c/
(2 S_f\sqrt{2Km_1 \vee m_2})\leq c/
(2 S_f\sqrt{2Km_1 })$. So
$$ I_3(\gamma) \geq \frac{1}{2}. $$
Next, we remark that
\begin{align*}
 \pi\left(  \left(U_{i,\ell}-U^0_{i,\ell}\right)^2 \leq \frac{c^2}{2m_1 r}
         \middle\vert\gamma\right)
 & \geq \int_{U^0_{i,\ell}}^{U^0_{i,\ell} + \frac{c}{\sqrt{2 m_1 r}}}
      \frac{1}{\gamma_j} f\left(\frac{u}{\gamma_j}\right){\rm d}u
      \\
 & \geq \int_{U^0_{i,\ell}}^{U^0_{i,\ell} + \frac{c}{\sqrt{2 m_1 r}}}
      \frac{\mathcal{C}_f}{\gamma_j} \widetilde{f}
                 \left(\frac{u}{\gamma_j}\right){\rm d}u.
\end{align*}
Remind that $1\leq \gamma_j \leq 2$ and $\tilde{f}$ is non-increasing so
\begin{align*}
 \pi\left(
        \left(U_{i,\ell}-U^0_{i,\ell}\right)^2 \leq \frac{c^2}{2m_1 r}
         \middle\vert\gamma\right)
 & \geq \frac{2 c\mathcal{C}_f}{\sqrt{2m_1 r}}
       \widetilde{f}\left(U^0_{i,\ell} + \frac{c}{\sqrt{2 m_1 r}}\right)
       \\
 & \geq \frac{2 c\mathcal{C}_f}{\sqrt{2m_1 r}}
       \widetilde{f}\left(U^0_{i,\ell} + 1  \right)
\end{align*}
as $c\leq 1 \leq \sqrt{m_1 r} $.
We plug this result and the lower-bound $I_3(\gamma) \geq 1/2$
into the expression of $I_1(\gamma)$ to get
$$
I_1(\gamma)
   \geq \frac{1}{2}
\left(\frac{2 c\mathcal{C}_f}{\sqrt{2m_1 r}}\right)^{m_1 r}
       \left[\prod_{\substack{1 \leq i \leq m_1\\1\leq \ell \leq r}} 
       \widetilde{f}\left(U^0_{i,\ell} + 1 \right)\right].
$$
Proceeding exactly in the same way,
$$
I_2(\gamma)
   \geq \frac{1}{2}
\left(\frac{2 c\mathcal{C}_f}{\sqrt{2m_2 r}}\right)^{m_2 r}
       \left[\prod_{\substack{1 \leq j \leq m_1\\1\leq \ell \leq r}} 
       \widetilde{f}\left(V^0_{j,\ell} + 1 \right)\right].
$$
So
\begin{multline*}
\int_E I_1(\gamma) I_2(\gamma) \pi(\gamma) {\rm d}\gamma \\
 \geq \int_E \frac{1}{4} \left(\frac{2 c\mathcal{C}_f}{\sqrt{2m_1 r}}\right)^{m_1 r}
  \left(\frac{2 c\mathcal{C}_f}{\sqrt{2m_2 r}}\right)^{m_2 r}
  \left[\prod_{\substack{1 \leq i \leq m_1\\1\leq \ell \leq r}} 
       \widetilde{f}\left(U^0_{i,\ell} + 1 \right)\right]
  \left[\prod_{\substack{1 \leq j \leq m_1\\1\leq \ell \leq r}} 
       \widetilde{f}\left(V^0_{j,\ell} + 1 \right)\right]
   \pi(\gamma) {\rm d}\gamma
   \\
 =  \frac{1}{4} \left(\frac{2 c\mathcal{C}_f}{\sqrt{2m_1 r}}\right)^{m_1 r}
  \left(\frac{2 c\mathcal{C}_f}{\sqrt{2m_2 r}}\right)^{m_2 r}
  \left[\prod_{\substack{1 \leq i \leq m_1\\1\leq \ell \leq r}} 
       \widetilde{f}\left(U^0_{i,\ell} + 1 \right)\right]
  \left[\prod_{\substack{1 \leq j \leq m_1\\1\leq \ell \leq r}} 
       \widetilde{f}\left(V^0_{j,\ell} + 1 \right)\right]
   \int_{E} \pi(\gamma) {\rm d}\gamma
\end{multline*}
and
\begin{align*}
\int_{E} \pi(\gamma) {\rm d}\gamma 
  & =
  \left(\int_{1}^2 h(x) {\rm d}x\right)^{r}
  \left(\int_{0}^{\frac{c}{2S_f\sqrt{2K m_1\vee m_2}}}
  h(x)
   {\rm d}x\right)^{K-r}
  \\
  & \geq
    \delta^{r} \alpha^{K-r} \left(\frac{c}{2S_f\sqrt{2K  m_1\vee m_2}}\right)^{\beta(K-r)}
   \\
  & \geq
   \delta^{r} \alpha^K \left(\frac{c}{2S_f\sqrt{2K  m_1\vee m_2}}\right)^{\beta K},
\end{align*}
using \autoref{condition_zero_h}.
We combine these inequalities, and we use trivia between $m_1$,
$m_2$, $m_1\vee m_2$ and $m_1 + m_2$ to obtain
 \begin{multline*}
 \mathcal{K}(\rho_{r,U^0,V^0,c},\pi)
 \leq
 2 (m_1 \vee m_2) r \log\left( \frac{\sqrt{2 (m_1\vee m_2) r}}
                        {c \mathcal{C}_f} \right)
 \\
 + \sum_{\substack{1 \leq i \leq m_1\\1\leq \ell \leq r}}
          \log\left(\frac{1}{ \widetilde{f}(U_{i\ell}^0 +1 )}
          \right)
 + \sum_{\substack{1 \leq j \leq m_2\\1\leq \ell \leq r}}
          \log\left(\frac{1}{ \widetilde{f}(V_{j\ell}^0 +1)}
          \right)
 \\
 + \beta K \log\left(\frac{2 S_f \sqrt{2  K (m_1 \vee m_2)}}
         {c}\right) + K\log\left(\frac{1}{\alpha}\right) + r\log\left(\frac{1}{\delta}\right) + \log(4).
 \end{multline*}
This ends the proof of the lemma.
\end{proof}

\subsection{Conclusion}
\label{conclusion}

We now plug \autoref{lemma_integral} and \autoref{lemma_kullback}
into \autoref{thm_dt}. We obtain, under
\autoref{condition_noise}, \autoref{condition_zero_h} and \autoref{condition_lowerb_f},
\begin{multline*}
 \E\left(
 \|\widehat{M}_{\lambda} - M\|_F^2
 \right)
 \leq
 \inf_{1 \leq r \leq K}\
 \inf_{(U^0,V^0)\in\mathcal{M}_r}\
 \inf_{0<c\leq \sqrt{Kr}}\
 \left\{
 \|U^0 V^{0\top}-M\|_F^2 \phantom{\sum^K}\right.
 \\
  +  \frac{2 (m_1 \vee m_2) r}{\lambda}
  \log\left( \frac{\sqrt{2  (m_1\vee m_2) r}}
                        {c \mathcal{C}_f} \right)
 \\
 + \frac{1}{\lambda}\sum_{\substack{1 \leq i \leq m_1\\1\leq \ell \leq r}}
          \log\left(\frac{1}{\widetilde{f}(U_{i\ell}^0 +1)}
          \right)
 + \frac{1}{\lambda}\sum_{\substack{1 \leq j \leq m_2\\1\leq \ell \leq r}}
          \log\left(\frac{1}{\widetilde{f}(V_{j\ell}^0 +1 )}
          \right)
 \\
 + \frac{\beta K}{\lambda} \log\left(\frac{2 S_f \sqrt{2  K (m_1 \vee m_2)}}
         {c}\right)
      + \frac{K}{\lambda}\log\left(\frac{1}{\alpha}\right)
            + \frac{r}{\lambda}\log\left(\frac{1}{\delta}\right)
         + \frac{1}{\lambda}\log(4)
  \\
\left.\phantom{\sum^K}    +  c \left( 1+ \|U^0\|_F + \| V^{0} \|_F+ 2 \|  U^0  V^{0\top} - M \|_F \right)^2 \right\}.
\end{multline*}
Remind that we fixed $\lambda = \frac{1}{4}$.
We finally choose
$$
c = \frac{1}{\left[1+ \|U^0\|_F + \| V^{0} \|_F+ 2 \|  U^0  V^{0\top} - M \|_F\right]^2}
$$
and so the condition $c \leq 1$ is always satisfied.
The inequality becomes
\begin{multline*}
 \E\left(
 \|\widehat{M}_{\lambda} - M\|_F^2
 \right)
 \leq
 \inf_{1 \leq r \leq K}\
 \inf_{(U^0,V^0)\in\mathcal{M}_r}\
 \left\{
 \|U^0 V^{0\top}-M\|_F^2 \phantom{\sum^K}\right.
 \\
  +  8(m_1 \vee m_2) r
   \log\left(
  \sqrt{2(m_1 \vee m_2)r}
  \right)
  \\
  +  8(m_1 \vee m_2) r
   \log\left(
   \frac{\left[1+ \|U^0\|_F + \| V^{0} \|_F+ 2 \|  U^0  V^{0\top} - M \|_F\right]^2}{ \mathcal{C}_f }
  \right)
 \\
 + 4\sum_{\substack{1 \leq i \leq m_1\\1\leq \ell \leq r}}
          \log\left(\frac{1}{\widetilde{f}(U_{i\ell}^0 +1)}
          \right)
 + 4\sum_{\substack{1 \leq j \leq m_2\\1\leq \ell \leq r}}
          \log\left(\frac{1}{\widetilde{f}(V_{j\ell}^0 +1)}
          \right)
 \\
 + 4\beta K    \log\left(
   \left[1+ \|U^0\|_F + \| V^{0} \|_F+ 2 \|  U^0  V^{0\top} - M \|_F\right]^2
  \right)
  \\
 + 4   \beta K \log\left(2 S_f \sqrt{2  K (m_1 \vee m_2)}\right)
   \\
\left. \phantom{\sum^K}     + r\left[4\log\left(\frac{1}{\delta}\right)\right] + 4 K \log\left(\frac{1}{\alpha}\right) + 4 \log(4) +1
 \right\},
\end{multline*}
which ends the proof.

\section*{Acknowledgements}

The authors are grateful to Jialia Mei and Yohann de Castro (Universit\'e
Paris-Sud) for insightful discussions and for providing many references on NMF,
and to the anonymous Referee for helpful comments.


\bibliographystyle{abbrvnat}
\bibliography{biblio}

\end{document}